# Inferencing the earth moving equipment-environment interaction in open pit mining

*M. Balamurali*

Research fellow, Australian Centre for Field Robotics, University of Sydney.
Email:mehala.balamurali@sydey.edu.au

## ABSTRACT

In mining, grade control generally focuses on blast hole sampling and the estimation of ore control block models with little or no attention given to how the materials are being excavated from the ground. In the process of loading trucks, the underlying variability of the individual bucket load will determine the variability of truck payload. Hence, accurate material movement demands a good knowledge of the excavation process and the buckets' interaction with the environment. However, equipment frequently goes into off-nominal states due to unexpected delays, disturbances or faults. The large amount of such disturbances causes information loss that reduces the statistical power and biases estimates, leading to increased uncertainty in the production. A reliable method that inferences the missing knowledge about the interaction between the machine and the environment from the available data sources, is vital to accurately model the material movement.

In this study, a two step method was implemented that performed unsupervised clustering and then predicted the missing information. The first method is Density-based spatial clustering of applications with noise (DBSCAN) (DBSCAN) based spatial clustering which divides the diggers' and buckets' positional data into connected loading segments. Clear patterns of segmented bucket dig positions were observed. The second model utilised Gaussian process regression which was trained with the clustered data and the model was then used to infer the mean locations of the test clusters. Bucket dig locations were then simulated at the inferred mean locations for different durations and compared against the known bucket dig locations. This method was tested at an open-pit mine in the Pilbara of Western Australia. The experimental results demonstrate the advantage of the proposed method in inferencing the missing information of bucket-environment interactions and therefore enables miners to continuously track the material movement.

## INTRODUCTION

Mining operations look for new approaches to maximize the ore and minimize the waste recovery at the process plant. The mining process includes drilling, blasting, loading and hauling and primary crushing. Having problems in one stage may lead to inefficiencies in the downstream process and therefore impact the production outcome. Among the mentioned steps, loading efficiency plays a vital role in increasing production and reducing costs as the loading equipment is the source of ore supply or waste removal (Balamurali 2022; Vasylchuk and Deutsch, 2015).

Hydraulic excavators are primarily used as loading equipment in large open pit mining operations around the world. In the process of loading trucks, the underlying variability of the individual bucket loads will determine the variability of truck payloads. Hence, accurate material movement demands a good knowledge of the excavation process and the buckets' interaction with the environment. The main hypothesis of this research is that these excavators can be used as a diagnostic tool to continuously and accurately estimate the moments (mean and variance) of the material loaded to each truck and therefore infer the propagation of material uncertainty down the material tracking pipeline. It can be achieved by knowing the time of engagement and disengagement of the bucket from the ground and its positional data coming from the sensor measurements. GPS sensors used on excavators facilitate collecting and analysing spatial and temporal information of the excavators and buckets during loading and dumping. Knowing the exact bucket dig locations helps the mining operators to infer the actual material taken from the ground by matching the grade control block models and the bucket dig locations.

However, equipment frequently goes into off-nominal states due to unexpected delays, disturbances or faults. A large amount of such disturbances causes loss of information about where the bucket interacts with the ground and thus reduces the accuracy in the estimation of the loaded material, leading to increased uncertainty in the production. In order to accurately model the material movement, it is crucial to have a reliable method that can infer the missing information regarding the bucket dig positions.

This paper proposes a two-step approach to infer the missing bucket dig locations and therefore help to estimate the moments of the material that goes on to trucks. Before carrying out further analysis on the data, splitting bucket dig positions into connected loading segments is necessary in data processing. This research proposes an unsupervised clustering method, DBSCAN, to group the connected bucket locations for each of the nearest excavator's GPS locations. The DBSCAN algorithm was originally developed by Ester et al. (1996) for clustering spatial points based on density difference.

The next step uses the Gaussian process regressor to learn the mapping between the excavator's GPS positions and the corresponding clusters' moments. The trained model is then used to infer the moments of missing clusters. Bucket dig locations are then simulated at missing clusters using the inferred moments of clusters. This twostep method was tested in a region at the Pilbara iron ore deposit situated in the Hamersley Province, Western Australia.

## METHODS

DBSCAN

A point is considered crowded if it has many other neighbouring points near it. The DBSCAN finds these dense points and places them and their neighbours in a cluster. The DBSCAN has two main parameters. The parameter ε (epsilon) defines the radius of a neighborhood with respect to some point. The next parameter 'MinPts' is the density threshold. If a neighbourhood includes the MinPts points, it will be considered as a dense region.

Consider a set of points in some space to be clustered. For the purpose of DBSCAN clustering, the points are classified as core points, (density-) reachable points and outliers, as follows:

- A point p is a core point if at least minPts points are within distance ε of it (including p).
- A point q is directly reachable from p if point q is within distance ε from core point p. Points are only said to be directly reachable from core points.
- A point q is reachable from p if there is a path $p_1, ..., p_n$ with $p_1 = p$ and $p_n = q$, where each $p_{i+1}$ is directly reachable from $p_i$. This implies that the initial point and all points on the path must be core points, with the possible exception of q.
- All points not reachable from any other point are outliers or noise points.

Now if p is a core point, then it forms a cluster together with all points (core or non-core) that are reachable from it. Each cluster contains at least one core point; non-core points can be part of a cluster, but they form its "edge", since they cannot be used to reach more points.

This clustering method does not need prior information of the number of clusters as input. The definitions and detailed parts of applying DBSCAN on GPS data can be found in Gong et al. (2015).

In this study, the input variables to the DBCAN are the x, y and z coordinates of buckets' dig and dump positions and x and y codinates of the excavator's GPS positions in local mins grid.

Gaussian Process Regression (GPR)

GPR is a generic supervised learning method designed to solve regression and probabilistic classification problems. The spread of the bucket locations within each cluster is estimated by calculating the mean and standard deviation of each cluster in both the x and y directions. The estimated moments are then used as target values for training the GPR. GPR learns the mapping between the excavator's GPS positions and the target values. The trained model is then used to infer the mean and standard deviation (in both x and y directions) of each unknown cluster using the given GPS positions of the excavator. The scikit-learn implementation for Gaussian Process Regressor was used in this study. Scikit-learn is an opensource package that provides various machine learning tools written in python (Pedregosa et al. 2011).

## RESULTS AND DISCUSSION
**DBSCAN to find the representative clusters**

DBSCAN is used to group bucket dig positions (x, y and z coordinates) into clusters (representative bucket dig locations) and noise due to their difference in spatial density (Figure1b). Figures 1c and 1d show the plan and vertical view of bucket dig locations after removing the outliers from Figure 1b. Each representative cluster is associated with its closest excavator's GPS positions (Figure 1a). Figure1 shows that well separated clusters are achieved in both horizontal and vertical planes using DBSCAN.

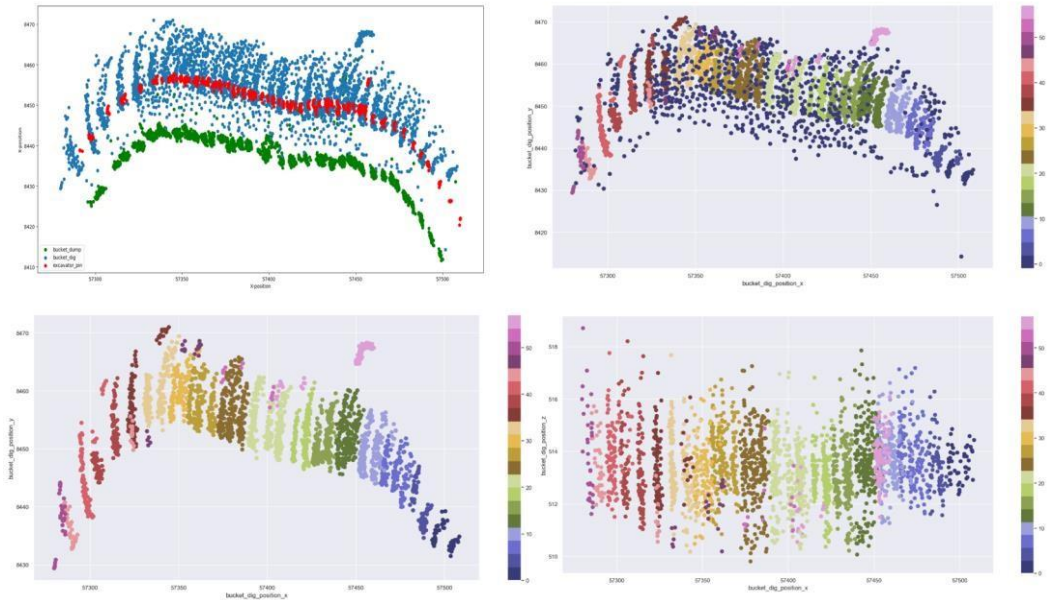

*Figure 1(a)shows the plan view of bucket dig locations, excavator's gps location and bucket dump locations in the test region. (b)shows the segmented bucket locations and noise data. Each segment is coloured according to the cluster labels. (c) and (d) show the clusters on plan and vertical view after removing the noise.*

**Inferencing the representative bucket dig location**

Scenario 1: Prediction on random location

In scenario 1, during the training process some of the moments of clusters were removed at random locations from the training samples and the model was trained with remaining samples. As can be seen in Figure 2a, the trained model was then used to infer the unknown moments at random locations in both the x and y directions using the excavator's GPS positions. The visual assessment in Figure2a shows that the inferred values are reasonably closer to the ground truth values.

Using the predicted mean and variance, gaussian samples were simulated at the missing locations in x and y directions and the bucket dig locations that fell in each removed cluster were recreated. Figure 3b shows the plan view of the Gaussian simulated bucket dig locations at the inferred mean clusters' locations and Figure 3c compares the simulated bucket dig locations with the ground truth bucket dig locations.

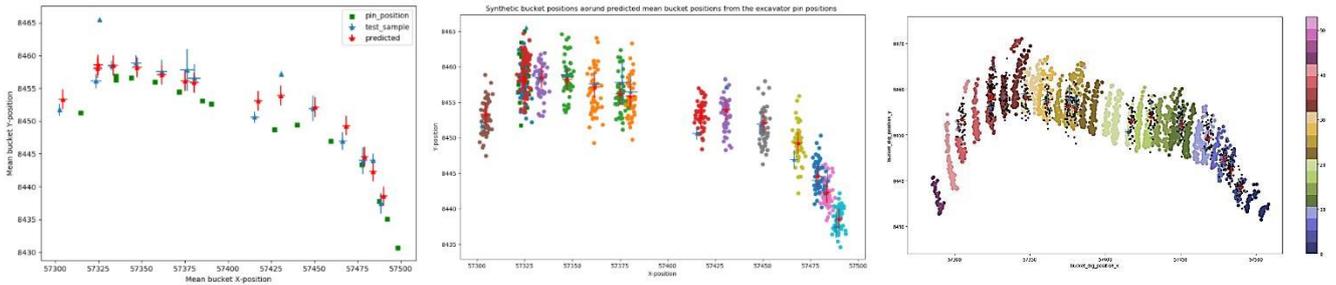

*Figure 2 shows the plan view of (a) ground truth and predicted moments of clustered bucket dig locations and excavators' pin position. (b) shows the simulated bucket locations and (c) compares the simulated locations(black) with recorded bucket locations (locations are coloured according to cluster labels).*

Scenario 2: Prediction on continuous locations for a longer period

Scenario 2 was implemented to evaluate the accuracy of the model prediction for a longer period. As can be seen in Figure 4a, a continuous bucket dig locations were removed for a longer period and the corresponding cluster moments were removed from the training samples. GPR was then trained on the remaining samples and the model predicted the moments on test locations.

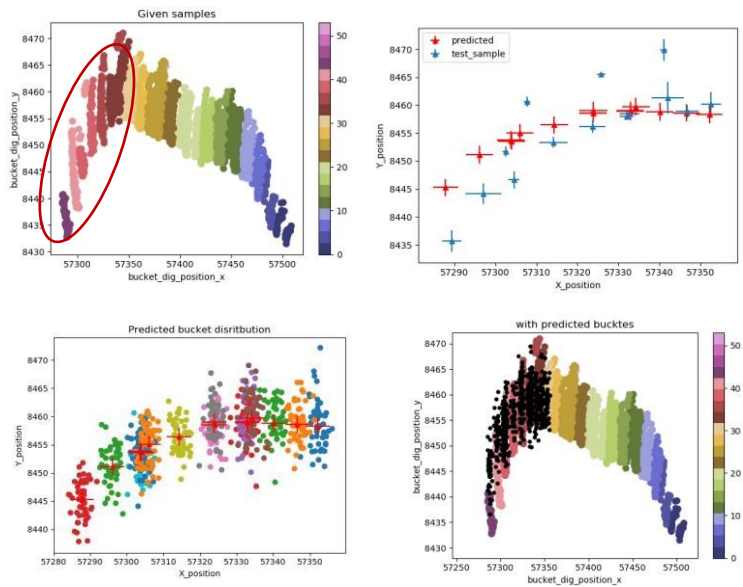

*Figure 3 shows the plan view of (a) given bucket dig locations coloured by its cluster labels*
*(b) shows the inferred moments of the test cluster samples and the ground*
*truth values in the circled area of (a). (c) shows the simulated bucket locations and (d)*
*compares the simulated locations(black) with given bucket locations (locations are*
*coloured according to cluster labels).*

Figure 3 compares the inferred clusters' moments on continuous timestamps and simulated bucket dig locations with the ground truth values. Figure 3b shows that the distance between the inferred means and the ground truth cluster means increases when the test locations are further away from the training locations. This might be due to the large time interval as the model has not been sufficiently trained to make predictions for longer periods or due to the steep change on the excavator's path where this kind of pattern hasn't been captured in the trained model. Further studies with longer sequence data at different regions and large excavators' GPS data during loading, can improve the model prediction accuracy. Even though the distance between the inferred and ground truth cluster means has increased with the steep curve of the excavator's travelling path (Figure 3b), the simulated bucket dig locations can still form clusters closer to the clusters previously obtained by DBSCAN.

This study shows that the accuracy of the model predictions is high on random locations when the model is trained on closer samples for shorter time periods compared to the predictions made for longer periods on continuous locations. However, in both circumstances, the simulated bucket locations form the clusters that are reasonably close to the clusters from the actual bucket dig locations.

Knowing the exact bucket dig locations helps the mining operators to infer the actual material taken from the ground by matching the grade control block models and the bucket dig locations. However, imperfect mining such as blasting, sheeting and grading of roads and benches will move ore and waste around the boundaries. Due to geometry, the operation of excavation equipment cannot be physically matched with the shape and size of the ore body, and therefore excavators blend the material further (Isaaks et al., 2014, Dimitrakopoulos & Godoy, 2014; Verly, 2005). Hence, the bucket dig locations are recorded in a way that makes it highly uncertain in knowing the exact material taken from the ground.

In this case, bucket dig locations simulated by the proposed model can still be used as bucket-material interaction points when a large number of bucket dig locations are missed and we need to know the distribution of the material loaded onto trucks from the corresponding locations. However, the uncertainty associated with the material taken from the inferred dig locations will be larger compared to the uncertainty of the loaded material when the dig locations are known.

The proposed model can be further improved by inferencing the representative load dig locations for each truck instead of inferencing large connected bucket dig locations (clusters) that correspond to the excavator's GPS positions. However, it would be more accurate if the model could infer each individual missing bucket dig location. This can be achieved by including more samples collected from larger study areas and other information coming from the sensors used in excavators and trucks. Using different spatial temporal clustering methods such as ST-DBSCAN and Hidden Semi Markov Model (HSMM), can improve the segmentation of bucket dig locations corresponding to each truck. Incorporating a time variable helps to identify sudden changes. Sudden increases can be observed when the excavator moves forward to the next load point or is waiting for the next truck. If such a "sudden increase" is found, the cluster can be divided into two clusters at the point of sudden increase. This information helps to further refine the cluster results. These are subject to further investigation.

**Conclusion**

A model approach is presented which can be used to infer the missing bucket dig locations when the bucket sensors frequently go into off-nominal states due to unexpected disturbances or faults. Therefore, the proposed study facilitates to continuously tracking the material at subsequent locations in the material tracking pipeline.

Several limitations and future research directions in the methodology have also been noted. To address this, the proposed model should be improved by evaluating other clustering methods by incorporating both temporal and spatial variables coming from different load and haul equipment at the time of mining. The GPR method used to regress the location information in this study should also be compared with other regressors to confirm its superiority; meanwhile, other possible sensor information can also be included in the regressor to improve the prediction accuracy.

## Acknowledgements

This work has been supported by the Australian Centre for Field Robotics and the Rio Tinto Centre for Mine Automation, the University of Sydney.